\documentclass[a4paper, conference]{IEEEtran}
\IEEEoverridecommandlockouts
\usepackage{abbrv}
\usepackage[dvipsnames]{xcolor}
\usepackage{url}
\usepackage{tikz}
\usepackage{comment}
\usepackage{amsthm}
\usepackage{amsmath,amssymb,bbm}
\usepackage{colortbl}
\usepackage{epsfig}
\usepackage{caption}
\usepackage{soul}
\usepackage{orcidlink}
\usepackage[utf8]{inputenc}
\usepackage{listings}
\usepackage[T1]{fontenc}
\usepackage{booktabs}
\usepackage{amsfonts}
\usepackage{nicefrac}
\usepackage{makecell}
\usepackage{microtype}
\usepackage{threeparttable}
\usepackage{algorithm-mine}
\usepackage{numprint}
\usepackage{siunitx}
\usepackage{etoolbox}
\usepackage{cancel}
\newcommand{\ubold}{\fontseries{b}\selectfont}
\robustify\ubold
\usepackage{wrapfig}
\usepackage{algpseudocode}
\usepackage[skip=2pt,font=footnotesize,labelfont=footnotesize]{subcaption}
\usepackage{pifont}
\usepackage{lineno}
\usepackage{array,multirow,graphicx}
\usepackage{tabularx}
\usepackage{listings}
\usepackage[para]{footmisc}
\usepackage{lettrine}

\usepackage[accsupp]{axessibility}
\usepackage{accsupp}
\usepackage{pgfplots}
\lstdefinelanguage{json}{
    basicstyle=\normalfont\ttfamily,
    numbers=left,
    numberstyle=\scriptsize,
    stepnumber=1,
    numbersep=8pt,
    showstringspaces=false,
    breaklines=true,
    frame=lines,
    backgroundcolor=\color{background},
    literate=
     *{0}{{{\color{numb}0}}}{1}
      {1}{{{\color{numb}1}}}{1}
      {2}{{{\color{numb}2}}}{1}
      {3}{{{\color{numb}3}}}{1}
      {4}{{{\color{numb}4}}}{1}
      {5}{{{\color{numb}5}}}{1}
      {6}{{{\color{numb}6}}}{1}
      {7}{{{\color{numb}7}}}{1}
      {8}{{{\color{numb}8}}}{1}
      {9}{{{\color{numb}9}}}{1}
      {:}{{{\color{punct}{:}}}}{1}
      {,}{{{\color{punct}{,}}}}{1}
      {\{}{{{\color{delim}{\{}}}}{1}
      {\}}{{{\color{delim}{\}}}}}{1}
      {[}{{{\color{delim}{[}}}}{1}
      {]}{{{\color{delim}{]}}}}{1},
}

\pgfplotsset{compat=1.16}

\definecolor{cvprblue}{rgb}{0.21,0.49,0.74}
\definecolor{keyword}{rgb}{.224,.451,.686}
\definecolor{tabhighlight}{HTML}{e5e5e5}
\definecolor{lightblue}{rgb}{0.63, 0.79, 0.95}
\definecolor{babypink}{rgb}{0.96, 0.76, 0.76}
\definecolor{skyblue}{rgb}{0.53, 0.81, 0.92}
\definecolor{wheat}{rgb}{0.96, 0.87, 0.7}
\definecolor{delim}{RGB}{20,105,176}
\colorlet{punct}{red!60!black}
\colorlet{numb}{magenta!60!black}

\def\BibTeX{{\rm B\kern-.05em{\sc i\kern-.025em b}\kern-.08em
    T\kern-.1667em\lower.7ex\hbox{E}\kern-.125emX}}
\hypersetup{colorlinks,%
citecolor=green,%
filecolor=%
linkcolor=%
urlcolor=green
}

\begin{document}
\title{SATURN: Autoregressive Image Generation \\ Guided by Scene Graphs}

\makeatletter
\newcommand{\linebreakand}{%
  \end{@IEEEauthorhalign}
  \hfill\mbox{}\par
  \mbox{}\hfill\begin{@IEEEauthorhalign}
}
\makeatother

\author{
  \IEEEauthorblockN{Thanh-Nhan Vo~\orcidlink{0009-0007-8403-1240}}
  \IEEEauthorblockA{\textit{University of Science, VNU-HCM} \\
  \textit{Vietnam National University, Ho Chi Minh City, Vietnam}\\
  vtnhan@selab.hcmus.edu.vn}
\\

  \IEEEauthorblockN{Tam V. Nguyen~\orcidlink{0000-0003-0236-7992}}
  \IEEEauthorblockA{\textit{University of Dayton}\\
  \textit{Ohio, USA} \\
  tnguyen1@udayton.edu}

  \and

  \IEEEauthorblockN{Trong-Thuan Nguyen~\orcidlink{0000-0001-7729-2927}}
  \IEEEauthorblockA{\textit{University of Science, VNU-HCM}\\
  \textit{Vietnam National University, Ho Chi Minh City, Vietnam}\\
  ntthuan@selab.hcmus.edu.vn}
\\
  \IEEEauthorblockN{Minh-Triet Tran~\orcidlink{0000-0003-3046-3041}}
  \IEEEauthorblockA{\textit{University of Science, VNU-HCM}\\
  \textit{Vietnam National University, Ho Chi Minh City, Vietnam} \\
  tmtriet@fit.hcmus.edu.vn}
}

\maketitle

\begin{abstract}
State-of-the-art text-to-image models can generate photorealistic images, yet they often struggle to accurately capture the layout and object relationships implied by complex prompts. Scene graphs provide the missing structural signal. However, prior graph-guided generators have typically relied on heavy GAN or diffusion architectures, which lag behind modern autoregressive pipelines in terms of speed and fidelity. In this work, we introduce \textbf{SATURN} (\textbf{\textit{S}}tructured \textbf{\textit{A}}rrangement of \textbf{\textit{T}}riplets for \textbf{\textit{U}}nified \textbf{\textit{R}}endering \textbf{\textit{N}}etworks), a lightweight, drop-in extension to VAR-CLIP that translates a scene graph into a salience-ordered token sequence, enabling a frozen CLIP–VQ-VAE backbone to parse the graph while fine-tuning only the VAR transformer. Empirically, on the Visual Genome dataset, SATURN reduces the FID score from 56.45\% to 21.62\% and raises the Inception Score from 16.03\% to 24.78\%, achieving gains that outperform SG2IM and SGDiff without requiring additional modules or multi-stage training. Notably, our qualitative results further confirm improvements in object count fidelity and spatial relation accuracy. 
\end{abstract}


\section{Introduction}\label{sec:intro}

Understanding and representing the complex structure of visual scenes remains a fundamental challenge in computer vision, as models must capture semantic entities and their spatial and relational dynamics. In particular, scene graphs provide an effective solution by encoding a scene as a set of \texttt{$\langle$subject, relation, object$\rangle$} triplets, thereby unifying semantics and spatial layout into a single, interpretable structure. Strong performance is reported on tasks including image captioning ~\cite{nguyen2021defense}, Visual Question Answering (VQA) ~\cite{qian2022scene}, and cross-model retrieval ~\cite{yoon2021image}. On the generative modeling front, the field has progressed from early approaches based on Generative Adversarial Networks (GANs)~\cite{van2017neural} to more advanced models such as diffusion models~\cite{yang2022diffusion} and autoregressive (AR) transformers. Early pixel-level autoregression methods, such as PixelCNN~\cite{van2016conditional} and PixelRNN~\cite{van2016pixel}, successfully captured fine-grained local details but incurred substantial computational costs. To address these challenges, Visual AutoRegressive (VAR)~\cite{VAR} predicts discrete VQ-VAE codes instead of raw pixel values, significantly improving computational efficiency. Building on this foundation, VAR-CLIP~\cite{zhang2024var} further enhances text-to-image generation by integrating VAR transformers with CLIP~\cite{radford2021learning} text embeddings~\cite{radford2021learning,zhang2024var}, achieving strong performance in aligning textual descriptions with high-quality visual outputs.

However, a text prompt alone often fails to fully specify the intended composition: generators may miscount objects, distort spatial layouts, or hallucinate relations due to the inherent under-specification of geometry in language~\cite{Liu2022Compositional}. In addition, existing scene graph-based generation methods, such as SG2IM~\cite{johnson2018image} and SGDiff~\cite{yang2022diffusion}, rely on GAN backbones or multi-stage diffusion pipelines, which lag behind modern VQ-VAE and CLIP-based methods in terms of visual fidelity and architectural simplicity.


To this end, we propose that an explicit graph structure can be incorporated into the current VAR-CLIP without modifying its large vision–language priors. By converting each scene graph triplet into a concise, well-ordered text sequence, the necessary semantic information can be embedded through CLIP, allowing the latent-token transformer to learn how to leverage this structured signal. Specifically, our proposed approach aims to incorporate the compositional precision of scene graphs with the efficiency of VAR-CLIP, thereby enabling controllable image generation and facilitating interactive image editing.

\noindent\textbf{Contributions.} First, we present SATURN (\textbf{\textit{S}}tructured \textbf{\textit{A}}rrangement of \textbf{\textit{T}}riplets for \textbf{\textit{U}}nified \textbf{\textit{R}}endering \textbf{\textit{N}}etworks), which translates a scene graph into an ordered token sequence and feeds it directly to the VAR-CLIP conditioning interface. Second, without adding any new modules or training stages, SATURN lowers FID from 56.45 $\rightarrow$ 21.62 and raises IS from 16.03 $\rightarrow$ 24.78 on Visual Genome~\cite{krishna2017visual}, outperforming SG2IM~\cite{johnson2018image} and SGDiff~\cite{yang2022diffusion}, while retaining competitive quality on COCO~\cite{lin2014microsoft} with EGTR-generated~\cite{im2024egtr} scene graphs.

\section{Related Work} \label{relatedworks}
\subsection{Generative Models for Image Synthesis}

Generative modeling for image synthesis pursues enhanced visual realism and precise control. Early autoregressive methods like PixelCNN~\cite{van2016conditional} and PixelRNN~\cite{van2016pixel} meticulously modeled distributions pixel-by-pixel, achieving detail but suffering from slow sequential generation. Transformer architectures~\cite{ parmar2018imagetransformer} were integrated to capture long-range dependencies, though computational costs remained significant. VAR~\cite{VAR} offered a more practical balance via a coarse-to-fine prediction strategy, improving speed.
Concurrently, VQ-VAE~\cite{van2017neural} pioneered discrete representation learning using quantized codebooks, effectively mitigating posterior collapse and enabling synergy with autoregressive priors. However, capturing both global structure and local details proved challenging. Hierarchical multi-scale VQ-VAE~\cite{razavi2019generating} improved this using multi-level architectures, enhancing detail and showing hierarchy's importance.

\subsection{Multimodal Integration in Generation}
Recent studies investigate multimodal information as conditional constraints to enhance control within generative models. Integrating diverse inputs enhances model capabilities and aligns with user intent. Among these modalities, text-guided generation is a particularly active area of research. This method leverages rich natural language semantics via robust text encoders such as BERT~\cite{devlin2019bert}, and CLIP~\cite{radford2021learning}. Consequently, text guidance is combined with various generative architectures, including autoregressive methods like VAR-CLIP~\cite{zhang2024var}, diffusion models like SORA~\cite{videoworldsimulators2024}, and GANs like StackGAN~\cite{zhang2017stackgan} and GigaGan~\cite{kang2023scaling}.
The CLIP model~\cite{radford2021learning} is especially effective for directly guiding the image generation process without requiring model retraining. Techniques that leverage its shared text-image embedding space, such as~\cite{crowson2022vqgan}, iteratively optimize image representations to align with textual descriptions, often using CLIP as a flexible zero-shot scoring function. Furthermore, CLIP embeddings are widely adopted as conditioning inputs in diffusion models, such as Stable Diffusion~\cite{rombach2021highresolution}. In these frameworks, the embeddings steer the iterative denoising process, ensuring the final generated image closely aligns semantically with the input text prompt. Integrating CLIP provides a powerful mechanism for bridging the gap between textual descriptions and visual synthesis.

\subsection{Scene-Graph-Based Image Generation}
Text-guided generation still struggles, particularly with complex descriptions involving intricate object relationships, often capturing only subsets of input data or producing distorted results. To address these restrictions, scene graphs offer structured representations detailing objects and their connections. Well-known studies such as SG2IM~\cite{johnson2018image},\cite{vo2020visual}, and SGDiff\cite{yang2022diffusion} leverage the scene graph to represent complex relations within generation models. These techniques yield promising results, generating images that better preserve scene graph details and avoid unrealistic distortions. Notably, SGDiff~\cite{yang2022diffusion} employs scene graph embeddings as direct conditioning signals for diffusion models, analogous to text conditioning in text-guided methods, enhancing structural adherence.
In addition, methods have been developed to exploit the power of scene graphs further. For example, EGTR~\cite{im2024egtr} presents a lightweight technique to extract scene graphs from images by reusing auxiliary information from object detectors. This method broadens the application scope of scene graph-based image generation models, enabling them to handle tasks such as scene graph-based image editing.
\begin{figure*}[!htbp]
    \centering
    \includegraphics[width=0.9\linewidth]{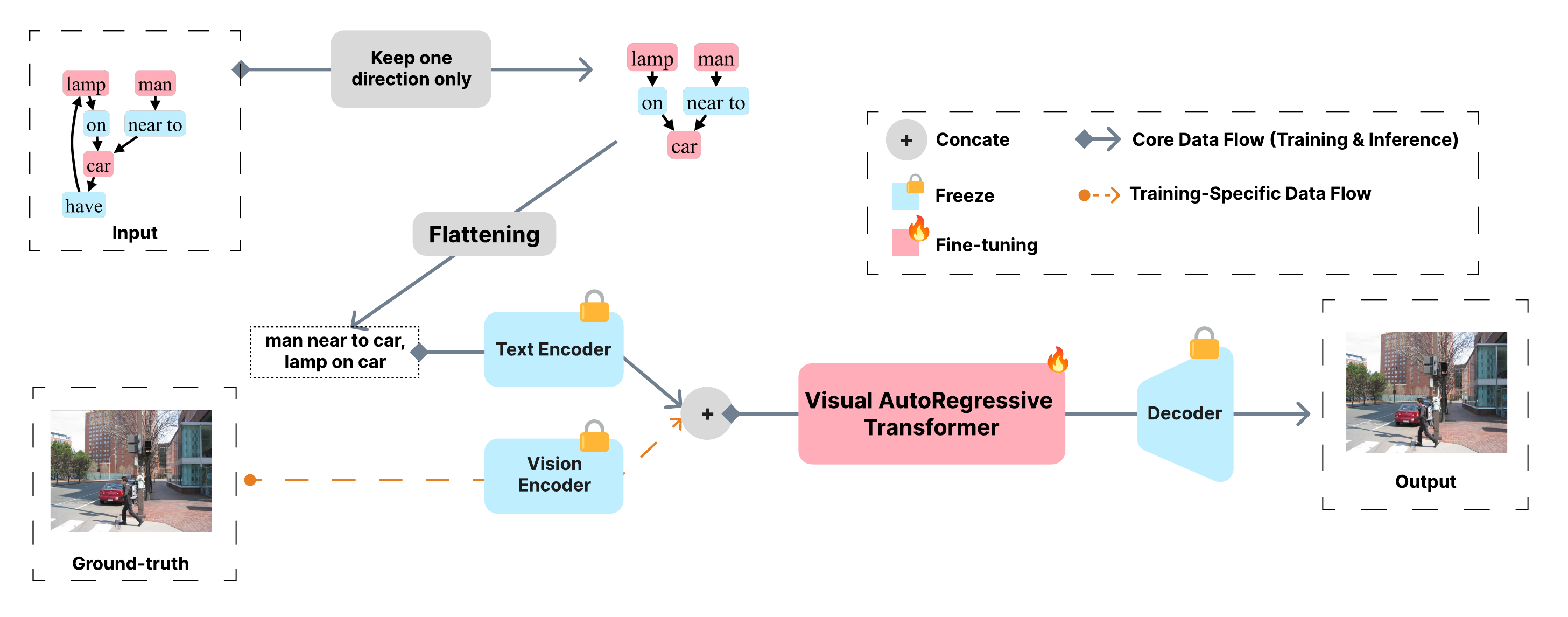}
    \caption{An illustration of our SATURN approach. A preprocessed scene graph is encoded into text embeddings (frozen text encoder). These are concatenated with visual embeddings (frozen vision encoder) to condition a fine-tuned VAR Transformer. The VAR generates visual representations, which a frozen decoder renders into the output image. Training utilizes ground-truth image encoding via the vision encoder (\textcolor{orange}{orange path}). \textbf{(Best viewed in color and with zoom.)}}
    \label{fig:method}
    \vspace{-5mm}
\end{figure*}
\subsection{Discussion}

\noindent\textbf{Limitations of Prior Work.} Text-conditioned generators~\cite{zhang2024var,videoworldsimulators2024,zhang2017stackgan} offer rich stylistic flexibility, but natural language is often imprecise regarding the number of objects, their spatial arrangement, and their interactions. As a result, these models frequently produce miscounts, misplaced objects, or hallucinated relations. In addition, scene graph-based methods~\cite{johnson2018image,yang2022diffusion} address this missing structure by explicitly encoding relational information; however, existing methods largely depend on GAN-based architectures or multi-stage diffusion models, which lag behind modern VQ-VAE and CLIP-based methods in both visual fidelity and efficiency.


\noindent\textbf{Advantages of Our Approach.} SATURN addresses these challenges by translating each scene graph into a compact, ordered token sequence that a frozen CLIP encoder can naturally interpret and then employing a single-stage VAR transformer to render the final image. This design preserves the fine-grained compositional control offered by scene graphs while leveraging the photorealism achieved by state-of-the-art text-to-image models without requiring additional modules, auxiliary losses, or multi-phase optimization. As a result, SATURN enables precise layout control, visual quality, and seamless integration into existing generative pipelines.

\section{Problem Formulation}\label{sec:prob_form}
In this work, we address the problem of generating images from scene graphs, where a scene graph is defined as a set of triplets, each augmented with bounding boxes and object categories. Formally, we define the scene graph as in Eqn.~\eqref{eq:graph}.
\begin{equation}
    \mathcal{G} = \{(s_i, r_i, o_i, b_i^s, b_i^o)\}_{i=1}^{N}, \qquad N = |\mathcal{G}|, \label{eq:graph}
\end{equation}
where $s_i$, $r_i$, and $o_i$ denote the subject, relation, and object of the $i$-th triplet, and $b_i^s$ and $b_i^o$ represent the corresponding bounding boxes of the subject and object, respectively.

Given a target image $I$, we mathematically formulate the task as a deterministic mapping from the scene graph $\mathcal{G}$ to the generated image $\hat{I}$, as expressed in Eqn.~\eqref{eq:e2e}.
\begin{equation}
    (\mathcal{G}, I_*) \xrightarrow{f_{\theta}} \hat{I}, \label{eq:e2e}
\end{equation}
where the parameters $\theta$ reside in the Visual AutoRegressive, which generates images conditioned on the scene graph using embeddings from frozen text and vision priors.

\section{Our Proposed Approach}\label{sec:approach}

Fig.~\ref{fig:method} illustrates the overall architecture of our proposed approach. In particular, the scene graph $\mathcal{G}$ is first converted into a caption $C$, which is then embedded using a frozen CLIP~\cite{radford2021learning} text encoder. The resulting text embedding conditions the VAR~\cite{VAR} to predict visual tokens, which are subsequently decoded into the image $\hat{I}$ by a frozen multi-scale VQ-VAE~\cite{van2017neural}.

\subsection{Scene Graph Encoding}\label{subsec:sg2cap}
To verbalize the structural information in the scene graph $\mathcal{G}$, each edge is first converted into a short textual phrase of the form $t_i = \textit{s}_i, \textit{r}_i, \textit{o}_i$, for $1 \le i \le N$. These phrases are then concatenated to form a raw caption $C_{\text{raw}} = t_1, t_2, \dots, t_N$.

In addition, we apply a two-step compression process to reduce token length. First, we remove bidirectional and duplicate relations (e.g., both $s_i$ related to $o_i$ and $o_i$ related to $s_i$), retaining only one direction per pair. Accordingly, the pruned set of phrases $C'$ is formally defined in Eqn.~\eqref{eq:prune}.
\begin{equation}
    C' = \{t_i \mid d_i = 0\}, \quad |C'| \le |C_{\text{raw}}|, \label{eq:prune}
\end{equation}
where $d_i$ is an indicator function used to identify and remove redundant relations. Next, we sort the remaining phrases based on the combined area of the subject and object bounding boxes, using the salience score $\alpha_i$, as formally defined in Eqn.~\eqref{eq:area}.
\begin{equation}
    \alpha_i = (w_i^s h_i^s) + (w_i^o h_i^o), \label{eq:area}
\end{equation}
where $(w, h)$ denote the width and height of the bounding boxes for the subject ($b_i^s$) and object ($b_i^o$). Based on this, the ordering prioritizes larger, more salient objects by placing them earlier in the caption. As a result, key scene details are better preserved during encoding, benefiting from CLIP’s early token bias. Ultimately, the ordered list of triplets forms the caption $C$, which serves as input for downstream image generation.


\subsection{Token-based Image Synthesis}\label{subsec:varclip}
A frozen multi-scale VQ-VAE encoder $\phi_v$ transforms an image $I \in \mathbb{R}^{H \times W \times 3}$ into a hierarchical grid of discrete visual tokens, $\mathbf{z} = \phi_v(I) \in {1, \dots, K}^L$, where $K = 8192$ and $L \approx 1024$ for images of size $H = W = 256$. This transformation converts the image from pixel space to token space, thereby enabling the application of language-style modeling techniques.

In addition, conditioned on the enhanced text embedding $\tilde{\mathbf{e}}_t$, the VAR Transformer $T_{\theta}$ learns to model the autoregressive distribution of visual tokens. Specifically, the transformer generates a sequence of tokens based on previously predicted tokens and the conditioning text embedding. This process is described by the autoregressive equation in Eqn.~\eqref{eq:trans}.

\begin{equation}
    p_{\theta}(\mathbf{z} \mid C) = \prod_{\ell=1}^{L} p_{\theta}(z_{\ell} \mid z_{<\ell}, \tilde{\mathbf{e}}_t), \label{eq:trans}
\end{equation}
where the probabilistic model learned by $T_{\theta}$, where $z_{\ell}$ are the visual tokens, and the conditional probability depends on the previous tokens ($z_{<\ell}$) and the enhanced text embedding $\tilde{\mathbf{e}}_t$.



\subsection{Loss Function}

To train the VAR model effectively, we adopt the standard training objective utilized in VAR-CLIP~\cite{zhang2024var}. Specifically, we optimize the negative log-likelihood (NLL) to guide the model in predicting visual tokens aligned with the semantics and spatial cues encoded in the textual description $\mathcal{C}$. This is formally defined in Eqn.~\eqref{eq:loss}.

\begin{equation}
\mathcal{L}(\theta) = -\mathbb{E}_{(\mathcal{C}, I_*)} \left[ \log p_{\theta}(\phi_v(I_*) \mid \mathcal{C}) \right]. \label{eq:loss}
\end{equation}

Here, $I$ represents the ground truth image corresponding to caption $\mathcal{C}$, and $\phi_v$ is the frozen VQ-VAE encoder. Eqn.~\eqref{eq:loss} encourages the VAR model, parameterized by $\theta$, to produce visual tokens that accurately reflect the content and layout described by the scene graph (via caption $\mathcal{C}$). 

During inference, the learned distribution $p_{\theta}(\cdot \mid \mathcal{C})$ is used to sample a sequence of tokens $\hat{\mathbf{z}}$. These tokens are then passed through the frozen VQ-VAE decoder $\psi_v$ to reconstruct the corresponding image $\hat{I}$, as formally defined in Eqn.~\eqref{eq:decode}.

\begin{equation}
\hat{I} = \psi_v(\hat{\mathbf{z}}). \label{eq:decode}
\end{equation}

The resulting image $\hat{I}$ is intended to faithfully reflect the structure and relations described by the conditioning scene graph $\mathcal{G}$, providing a visual representation that closely aligns with the semantic and spatial details encoded in the caption $\mathcal{C}$.

\section{Experiment Results}\label{sec:experiment}

\subsection{Implementation Details}
\noindent\textbf{Model Configuration.} Our experiments build upon a pre-trained VAR-CLIP model. We initialize our model using a publicly available VAR-CLIP checkpoint based on the VAR-d16 architecture \cite{zhang2024var}, which itself utilizes publicly available CLIP and VQ-VAE checkpoints. For fine-tuning, conducted on approximately 50,000 Visual Genome images for 50 epochs, we adopt a targeted approach: the weights of the integrated CLIP text encoder and the multi-scale VQ-VAE (both encoder and decoder) are kept frozen. Only the VAR transformer component itself is fine-tuned to adapt to our specific task of generating images from scene graph-derived captions. Training is conducted on the A100 GPU using the \texttt{Adam} optimizer with an initial learning rate of $3 \times 10^{-4}$, $(\beta_1, \beta_2) = (0.9, 0.95)$, cosine learning rate decay, and gradient clipping set to 1.0.

\noindent\textbf{Datasets.} Our proposed approach is trained on a subset of approximately 50,000 images from the Visual Genome~\cite{krishna2017visual} dataset, which offers detailed scene graph annotations. Evaluation is performed on a separate hold-out set of around 5,000 VG images. For comparative analysis against the VAR-CLIP baseline, we further evaluate our model on an additional set of approximately 5,000 images from the COCO~\cite{lin2014microsoft}. As COCO does not include native scene graph annotations, we synthetically generate them for the evaluation set using a pre-trained EGTR~\cite{im2024egtr}, a state-of-the-art scene graph generation.

\noindent\textbf{Metrics.} We evaluate the quality and semantic alignment of our generated images using several standard metrics. F\textit{réchet Inception Distance (FID)}~\cite{heusel2017gans} compares the statistical properties (mean and covariance) of features extracted from real and generated images using the Inception-v3 network. Lower FID scores indicate higher visual realism. \textit{Inception Score (IS)}~\cite{salimans2016improved} is used to assess the clarity and diversity of generated samples, where higher scores reflect outputs that are both recognizable and varied. Additionally, to measure semantic consistency with the input captions derived from scene graphs, we compute \textit{CLIP-based similarity scores}~\cite{radford2021learning}, using the cosine similarity between CLIP embeddings of the image and its corresponding text; higher scores indicate stronger alignment between visual content and textual description.




\subsection{Quantitative Results}\label{subsec:quant}

\begin{table}[!b]
  \centering
  \renewcommand{\arraystretch}{1.2}
  \caption{Comparison (\%) of baseline CLIP similarity for vanilla VAR-CLIP, measured using CLIP-L/14. \textit{Best results in \textbf{bold}.}}
  \begin{tabular}{lccc}
    \toprule
    Dataset & Mean & Max & Min \\ \midrule
    VG (no limit) & 18.64 & 29.50 & \textbf{5.34} \\
    COCO          & \textbf{22.56} & \textbf{36.41} & 3.07 \\ \bottomrule
  \end{tabular}
  \label{tab:VG_CLIP2}
  \vspace{-2mm}
\end{table}
Table~\ref{tab:VG_CLIP2} indicates that vanilla VAR-CLIP achieves a higher mean CLIP similarity on COCO (22.56\%) than on Visual Genome (18.64\%), despite employing identical model weights. Specifically, this discrepancy arises from the fact that COCO captions, expressed in natural language, more closely align with CLIP’s embedding space. In contrast, Visual Genome scene-graph captions, often generated through naïve triplet concatenation, deviate from CLIP’s linguistic priors, leading to markedly lower worst-case scores (minimum = 5.34\%). This baseline disparity highlights the need for a structure-aware conditioning scheme to model complex scene graphs.

\begin{table}[!t]
  \centering
  \renewcommand{\arraystretch}{1.2}
  \caption{Effect of relation count on vanilla VAR-CLIP (Visual Genome, CLIP-B/32). \textit{Best results in \textbf{bold}.}}
  \begin{tabular}{lccc}
    \toprule
    \# Relations & Mean & Max & Min \\ \midrule
    $\le$2      & \textbf{25.45} & \textbf{35.06} & \textbf{13.36} \\
    $\le$5      & 24.34 & 34.38 & 12.59 \\
    Unlimited   & 23.90 & 34.97 & 10.98 \\ \bottomrule
  \end{tabular}
  \label{tab:VG_CLIP}
  \vspace{-2mm}
\end{table}
In addition, Table~\ref{tab:VG_CLIP} examines this need more directly by varying the number of relations supplied to the vanilla VAR-CLIP model. Specifically, the results reveal an approximately linear decline in similarity: restricting captions to two relations yields a mean CLIP similarity of 25.45\%, whereas incorporating all relations reduces it to 23.90\%. Moreover, this pattern indicates that appending additional triplets pushes salient objects toward the end of CLIP’s 77-token window, where the transformer's attention weakens, resulting in semantic dilution. These findings further support SATURN’s pruning and salience-ordering strategies, which prioritize critical triplets early in the sequence to maintain fidelity in dense graphs.
\begin{figure}[!b]
\vspace{-2mm}
    \centering
    \includegraphics[width=\linewidth]{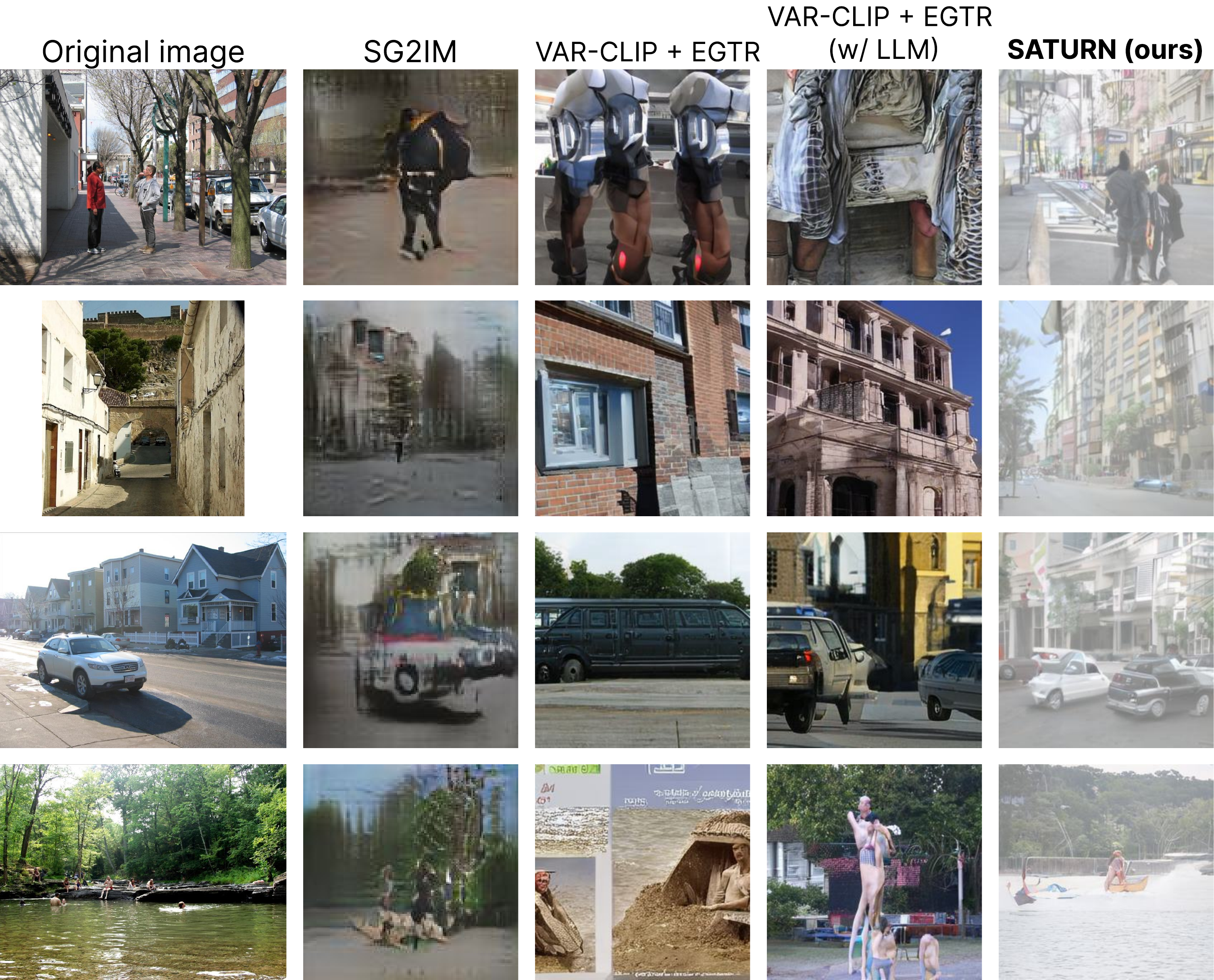}
    \caption{Visual comparison of image generation results from our SATURN approach and prior works. \textbf{(Best viewed in color and with zoom.)}}
    \vspace{-2mm}
    \label{fig:compare_othermodel}
\end{figure}

\begin{table}[!t]
  \centering
  \setlength{\tabcolsep}{4pt}
  \renewcommand{\arraystretch}{1.2}
  \caption{Comparison (\%) of SATURN and baselines on Visual Genome and COCO (lower FID, higher IS and CLIP scores are better). † COCO results use graphs predicted by EGTR; ‡ CLIP score is computed against graph-derived captions. \textit{Best results in \textbf{bold}.}}
  \begin{tabular}{l|ccc|cc}
    \toprule
    \multirow{2}{*}{Model} & \multicolumn{3}{c|}{VG} & \multicolumn{2}{c}{COCO} \\ \cline{2-6}
       & FID$\downarrow$ & IS$\uparrow$ & CLIP$\uparrow$ & IS$\uparrow$ & CLIP$\uparrow$ \\ \midrule
    SG2IM                 &  90.5   &  5.5 &  --   &  6.7 &  -- \\
    SGDiff                & 26.0    & 16.4 & -- & 17.8 & --\\
    VAR-CLIP (baseline)   & 56.45 & 16.03 & 18.64 & \textbf{34.54} & \textbf{22.56} \\
    \textbf{SATURN}       & \textbf{21.62} & \textbf{24.78} & \textbf{21.25}$^{\ddagger}$ & 15.41$^{\dag}$ & 20.98$^{\dag\ddagger}$ \\ \bottomrule
  \end{tabular}
  \label{tab:var-clip-finetune}
  \vspace{-2mm}
\end{table}

\begin{figure*}[!htbp]
    \centering
    \includegraphics[width=\linewidth]{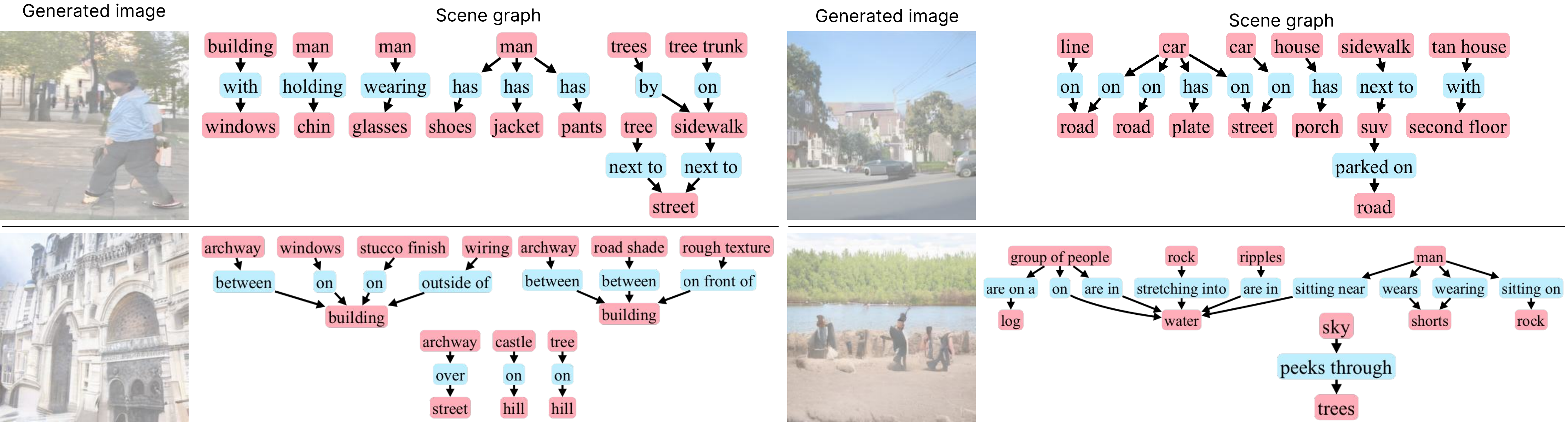}
    \caption{Example of scene graph conditioned image generation from our SATURN approach. \textbf{(Best viewed in color and with zoom.)}}
    \label{fig:imagewithsg}
    \vspace{-2mm}
\end{figure*}
The benefits of these improvements are reflected in Table~\ref{tab:var-clip-finetune}. After applying SATURN, the FID score drops significantly from 56.45\% to 21.62\%. In contrast, the IS score increases from 16.03\% to 24.78\% on the Visual Genome dataset, substantial gains that surpass those achieved by SG2IM or SGDiff, and accomplished without requiring additional modules or multi-stage training. The CLIP similarity score also rises to 21.25\%, indicating improved graph-level faithfulness. On the COCO dataset, SATURN remains competitive despite relying on EGTR-predicted graphs; the modest decrease in IS (34.54 $\rightarrow$ 15.41) is primarily attributable to errors introduced by the synthetic graphs, suggesting that simple graph-quality filtering could further narrow the remaining performance gap.

\subsection{Qualitative Results}\label{subsec:qual}
Beyond quantitative metrics, we conduct a qualitative analysis to assess the capabilities of our fine-tuned model visually. In particular, Fig.~\ref{fig:compare_othermodel} presents a visual comparison between images generated by SATURN and those produced by relevant prior works. The results indicate that SATURN achieves superior alignment with complex relational descriptions, demonstrating its effectiveness in capturing fine-grained structural details.

To directly illustrate the influence of input scene graph structure on the generated output, Fig.~\ref{fig:imagewithsg} presents representative pairs of input scene graphs and their corresponding images generated by our model conditioned on those graphs. This visualization enables direct inspection of how the model interprets and renders the provided structural information.

In addition, Fig.~\ref{fig:imagewithsg} presents a representative input scene graph alongside its corresponding generated image output to visualize the fundamental scene graph-to-image mapping. Building upon this single example, Fig.~\ref{fig:gallery} demonstrates the model's broader generative capabilities through a diverse set of 36 generated images. These examples, selected to cover various scene graph complexities and content types, illustrate the model's ability to produce visually coherent and semantically relevant outputs consistent with the provided structural descriptions.
\begin{figure}[!htbp]
    \centering
    \includegraphics[width=\linewidth]{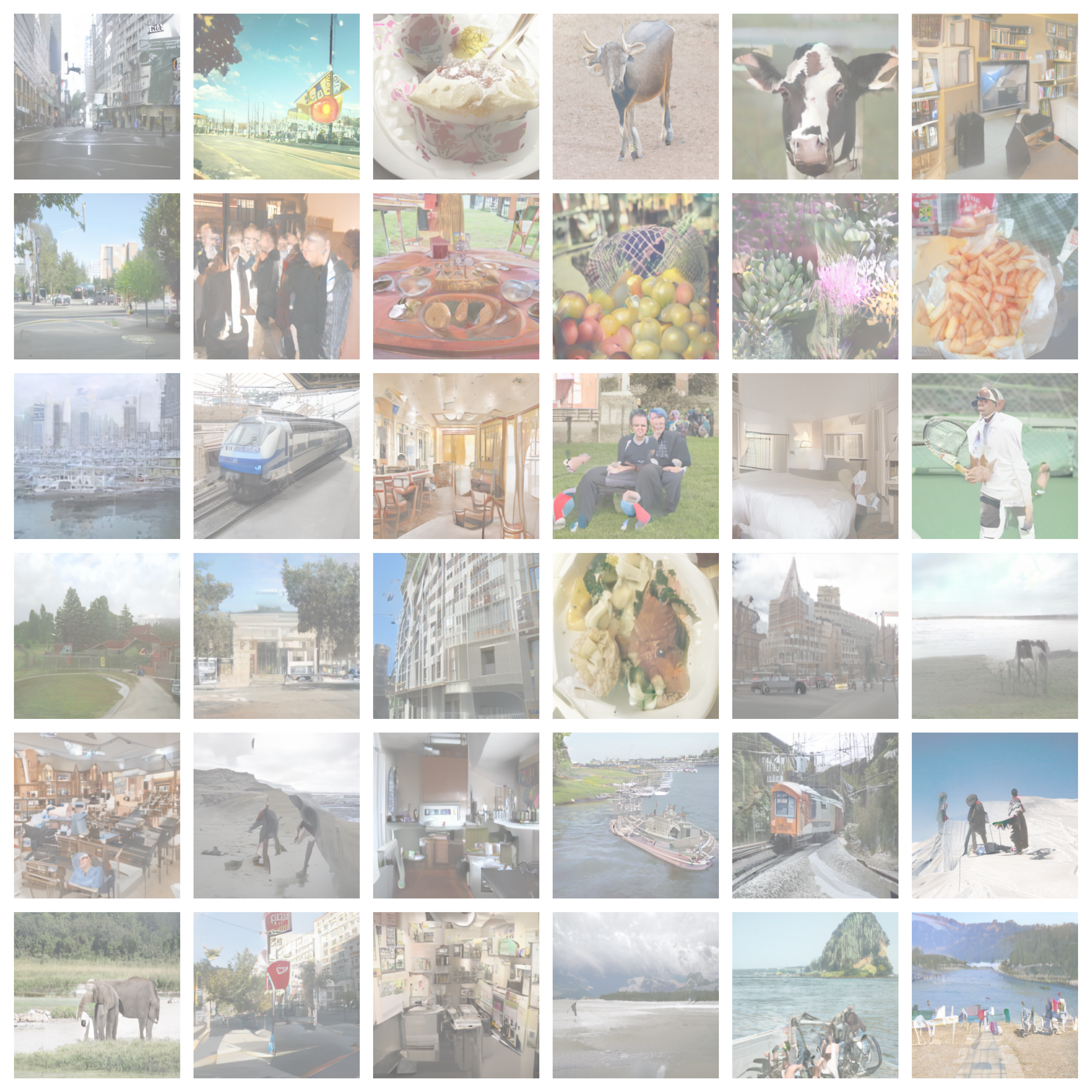}
    \caption{Examples of images generated by our SATURN approach. \textbf{(Best viewed in color and with zoom.)}}
    \label{fig:gallery}
    \vspace{-5mm}
\end{figure}

Moreover, we involve modifications to the input scene graphs to investigate the model's sensitivity and responsiveness to changes in the input structure. This process is illustrated in Fig.~\ref{fig:SGedit}, where we applied several editing operations to an initial scene graph, including changing existing objects, adding new relations, and deleting objects. Furthermore, we tested the model's response to semantically implausible modifications, such as inserting relations like ``\textit{group of people are in sky}''.
Typically, the resulting images demonstrate reasonable alignment with the respective modified scene graph descriptions, indicating the model attempts to follow the altered structural input. However, we observed that images generated from different modifications originating from the same initial scene often lack strong visual coherence or similarity among themselves. This suggests that while the model responds to specific edits, it may generate substantially different global scene layouts even for relatively localized changes to the input graph structure.

\begin{figure*}[!htbp]
    \centering
    \includegraphics[width=0.95\linewidth]{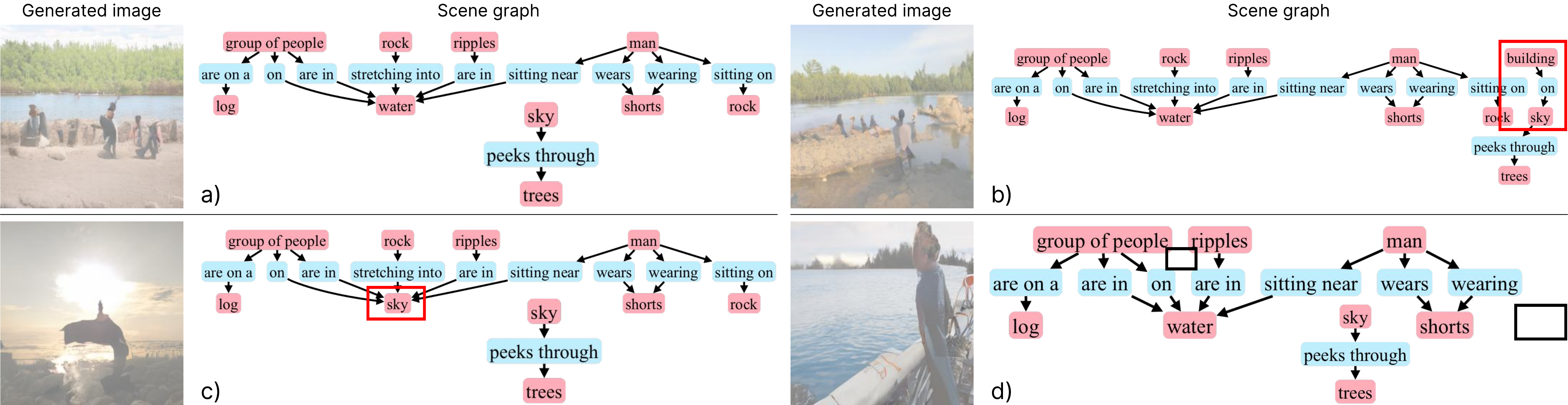}
    \caption{Scene graph editing results: a) Original graph. b) New "building" object and its relation with "sky" added. c) ``Water'' object replaced by ``Sky''. d) Two ``rock'' objects and their associated relations removed. \textbf{(Best viewed in color and with zoom.)}}
    \label{fig:SGedit}
    \vspace{-5mm}
\end{figure*}
\section{Conclusion}\label{sec:conclusion}


In this work, we introduced the SATURN approach, which translates scene graphs into a salience-ordered token sequence, bridging the gap between graph-level structure and state-of-the-art text-to-image generation. SATURN reduces the FID score and increases the IS score, outperforming heavyweight graph-guided baselines without additional stages or modules. Additionally, qualitative analyses confirm accurate object counts, faithful layouts, and editable compositional control.

Despite these positive outcomes, our experiments also revealed key challenges and limitations. The model’s sensitivity to the quality of synthetically generated scene graphs and occasional inconsistencies in maintaining global layout during editing highlight important areas for further improvement. While we validate the potential of integrating explicit structural information from scene graphs into text-conditioned generative models in this work, it also underscores the need for more robust methods for graph translation and greater resilience to variations in input graph quality. Therefore, future work could explore graph encoding techniques to enhance controllable generation under diverse conditions.

\textbf{Acknowledgment} This research is supported by research funding from Faculty of Information Technology, University of Science, Vietnam National University Ho Chi Minh City.
{\small
\bibliographystyle{ieeetr}
\bibliography{main}
}
\end{document}